\documentclass[10pt,sigconf]{acmart}

\usepackage{booktabs} 

\usepackage{balance}
\usepackage{tabularx}
\usepackage{tabularx}
\usepackage{multirow}
\usepackage{diagbox}
\usepackage{color}

\graphicspath{{figure/}{figures/}}

\setcopyright{rightsretained}

\acmDOI{10.475/123_4}

\acmISBN{123-4567-24-567/08/06}

\acmConference[WSDM'20]{International Conference on Web Search and Data Mining}{February 2020}{Houston, Texas, USA} 
\acmYear{2017}
\copyrightyear{2017}

\acmPrice{15.00}

\begin{document}
\title{Biomedical Evidence Generation Engine}
\titlenote{Produces the permission block, and copyright information}
\subtitle{Extended Abstract}

\author{Sendong Zhao}
\orcid{1234-5678-9012}
\affiliation{%
  \institution{Weill Medical College, Cornell University}
  \streetaddress{425 E 61st St}
  \city{New York} 
  \state{NY} 
  \postcode{10065}
}
\email{sez4001@med.cornell.edu}

\author{Fei Wang}
\orcid{1234-5678-9012}
\affiliation{%
  \institution{Weill Medical College, Cornell University}
  \streetaddress{425 E 61st St}
  \city{New York} 
  \state{NY} 
  \postcode{10065}
}
\email{few2001@med.cornell.edu}

\renewcommand{\shortauthors}{Zhao and Wang}

\begin{abstract}
With the rapid development of precision medicine, a large amount of health data (such as electronic health records, gene sequencing, medical images, etc.) has been produced. It encourages more and more interest in data-driven insight discovery from these data. It is a reasonable way to verify the derived insights in biomedical literature. However, manual verification is inefficient and not scalable. Therefore, an intelligent technique is necessary to solve this problem. In this paper, we propose a task of biomedical evidence generation, which is very novel and different from existing NLP tasks. Furthermore, we developed a biomedical evidence generation engine for this task with the pipeline of three components which are a literature retrieval module, a skeleton information identification module, and a text summarization module.
\end{abstract}

\ccsdesc[500]{Information systems~Information retrieval}
\ccsdesc[500]{Information systems~Information extraction}
\ccsdesc[500]{Computing methodologies~Natural language processing}
\ccsdesc[500]{Applied computing~Health informatics}


\keywords{Biomedical literature mining, Document retrieval, Information extraction, Text summarization, Biomedical evidence}

%

\maketitle

\section{Introduction}
With the arrival of precision medicine era, more and more health data (such as electronic health records, gene sequencing, medical images, etc.) are becoming readily available. Data-driven insight discovery from these data is becoming popular and crucial. One key component in such data-driven analytics pipeline is the verification of the derived insights in published biomedical articles (e.g., from PubMed), which currently is a manual process and not scalable.

In this paper, we propose the task of biomedical evidence generation. The goal of this task is to summarize key points in the biomedical publication in accordance with the query derived from the data-driven discoveries. This task is very novel and different from existing NLP tasks like general domain text summarization, relation extraction, and reading comprehension. Specifically, it has its own characteristics: 1) the generated evidence must be consistent facts according to the original publication; 2) the generated evidence must be in response to the query; 3) the generated evidence must be expressed in sentence(s); 4) the query derived from the data-driven discoveries
is usually entities (such as `diabetes' and `metformin') rather than questions in natural language.
1) and 2) distinguish this task from general domain text summarization, 2) and 3) distinguish it from relation extraction, 3) and 4) distinguish it from reading comprehension. 
Table~\ref{tab:exp} presents an example to illustrate our proposed biomedical evidence generation task. It is very clear that this task is an integrated task of document retrieval, information extraction, and text summarization and much more difficult than any of them.

\begin{table*}[htb]
\centering
\small
\caption{An example to illustrate our proposed biomedical evidence generation task.}
\begin{tabularx}{\textwidth}{|X|X|X|X|}
\hline
\textbf{Query} & \textbf{Text in literature} & \textbf{Skeleton information} &\textbf{Summary}\\
\hline
diabetes, metformin &....... Lifestyle changes and \textit{treatment} with \textbf{metformin} both \textit{reduced the incidence of} \textbf{diabetes} in persons at high risk. The lifestyle \textit{intervention} was more effective than \textbf{metformin} ......& treatment, metformin, reduced the incidence of, diabetes, intervention & \textbf{metformin} treatment prevent \textbf{diabetes}, but lifestyle intervention is more effective\\
\hline
\end{tabularx}
\label{tab:exp}
\end{table*}

\begin{figure*}[tb]
    \centering
    \includegraphics[width=\textwidth]{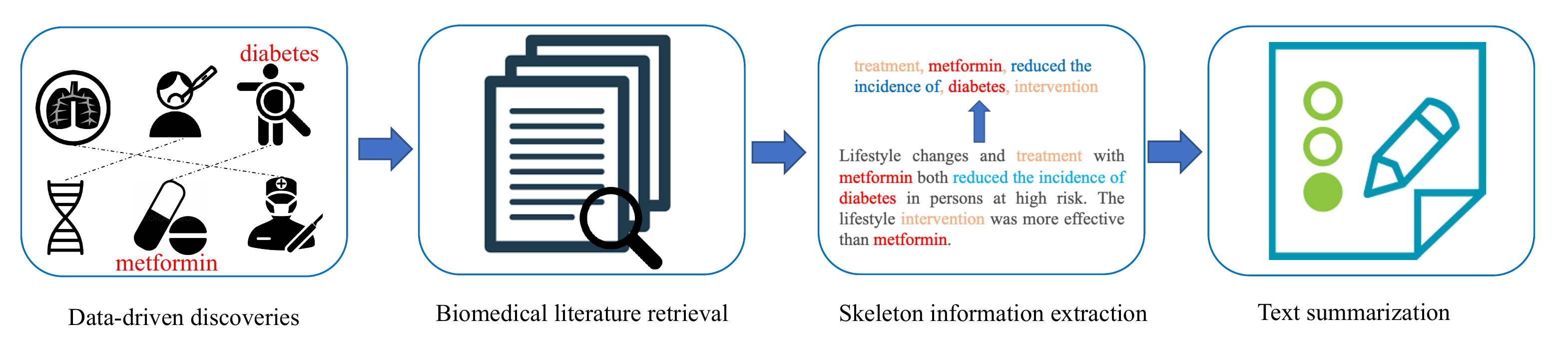}
    \caption{Pipeline of our system to generate biomedical evidence.}
    \label{fig:framework}
\end{figure*}

To solve this task, we developed a biomedical evidence generation engine (BEGE), which includes three main modules: 1) a literature retrieval module which can retrieve publications that are highly relevant to the query derived from the data-driven discoveries; 2) a skeleton information identification module extract special query-based words or expression (call it skeleton) to make sure they are in accordance with the query.
3) a text summarization module summarizes the key information from the retrieved articles with the key words/expressions provided as the skeleton

In the biomedical literature module, the query should be expanded due to diverse forms of biomedical entities. To this end, we expand different types of entities with different biomedical knowledge bases. In particular, we expand disease names with their different alias and abbreviations with Lexigram \footnote{https://www.lexigram.io/}, expand gene names with their alias in NCBI \cite{sherry2001dbsnp}, and expanse drugs with drug knowledge bases \cite{zhu2018drug}, etc. Then the expanded query is fed to the conventional retrieval model (such as BM25 \cite{robertson2009probabilistic}) or deep learning models \cite{zhao2019graphene,zhao2019interactive}. A list of relevant publications to the query is presented as the output of this module.

In the skeleton information identification module, the similarity measure is conducted to select relevant words or expressions matching the query in each retrieved article. The identified relevant words or expression might be synonymous to entities in query, or most related entities with query like ``diabetes" and ``metformin", or triggers/indicators to expose a claim or finding, such as ``is linked to", ``increase the risk of" and ``demonstrate", etc.

In the text summarization module, the summary of the evidence is presented in natural language text for each article. The skeleton information should define the main structure of the summarization. Meanwhile, it should have the same meaning as the original corresponding text.
Therefore, we propose to use a skeleton-based text summarization model, which takes skeleton information as strong attention and generates the shortening rephrase of key piece of text from biomedical literature.

Figure~\ref{fig:framework} presents the processing pipeline of our system. The query is the discovery derived from co-occurrence in biomedical data obtained through data-driven methods. The query after expansion would be fed to a retrieval model to get relevant publications which contain all entities in query or their corresponding alias. A skeleton information extraction is applied to get key information in response to the query. As shown in the skeleton information extraction part of the Figure\ref{fig:framework}, the exact matched words with the query are marked in red, words with similar semantics with the query are marked in orange, and trigger words for relations are marked in blue.
The extracted skeleton information is provided to text summarization to make sure the generated summary precisely responds to the query and rephrases the meaning of the original text.

\bibliographystyle{acm}
\bibliography{sigproc} 

\end{document}